\documentclass[journal,twoside,web]{ieeecolor}
\usepackage{jsen}
\usepackage{amsmath,amssymb,amsfonts}
\usepackage{algorithmic}
\usepackage{graphicx}
\usepackage{textcomp}
\usepackage{wrapfig}
\usepackage[english]{babel}
\usepackage[backend=bibtex,style=ieee,sorting=none,isbn=false,url=false,doi=false]{biblatex}
\def\BibTeX{{\rm B\kern-.05em{\sc i\kern-.025em b}\kern-.08em
    T\kern-.1667em\lower.7ex\hbox{E}\kern-.125emX}}
\definecolor{abstractbg}{rgb}{0.89804,0.94510,0.83137}
\begin{document}
\title{GTac: A Biomimetic Tactile Sensor with Skin-like Heterogeneous Force Feedback for Robots}
\author{Zeyu Lu, \IEEEmembership{Student Member, IEEE}, Xingyu Gao, and Haoyong Yu, \IEEEmembership{Senior Member, IEEE}
\thanks{This work was supported by Agency for Science, Technology and Research, Singapore, under the National Robotics Program, with A*star SERC Grant No.: 192 25 00054.}
\thanks{Zeyu Lu is with Department of Biomedical Engineering, National University of Singapore, Singapore 117575 (e-mail: zeyu.lu@u.nus.edu).}
\thanks{Xingyu Gao is with School of Mechanical and Electrical Engineering, Guilin University of Electronic Technology, Guilin, Guangxi 541004, China (e-mail: gxy1981@guet.edu.cn).}
\thanks{Haoyong Yu is with Department of Biomedical Engineering, National University of Singapore, Singapore 117575 (e-mail: bieyhy@nus.edu.sg).}
\thanks{This work has been submitted to the IEEE for possible publication. Copyright may be transferred without notice, after which this version may no longer be accessible.}
\thanks{© 20xx IEEE. Personal use of this material is permitted. Permission from IEEE must be obtained for all other uses, in any current or future media, including reprinting/republishing this material for advertising or promotional purposes, creating new collective works, for resale or redistribution to servers or lists, or reuse of any copyrighted component of this work in other works.}
}

\IEEEtitleabstractindextext{%
\fcolorbox{abstractbg}{abstractbg}{%
\begin{minipage}{\textwidth}%
\begin{wrapfigure}[16]{r}{3.3in}%
\includegraphics[width=3.2in]{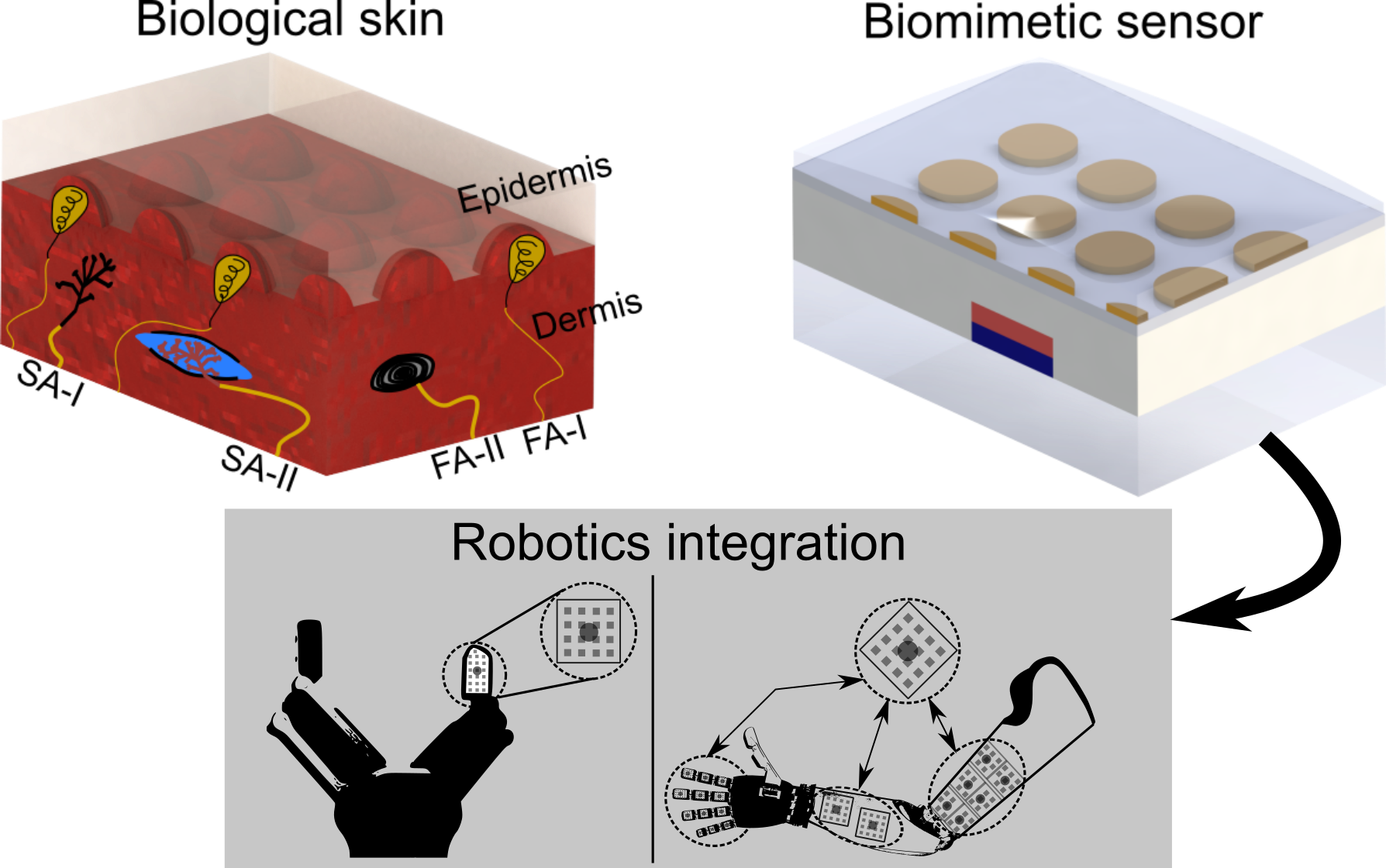}%
\end{wrapfigure}%
\begin{abstract}
The tactile sensing capabilities of human hands are essential in performing daily activities. Simultaneously perceiving normal and shear forces via the mechanoreceptors integrated into the hands enables humans to achieve daily tasks like grasping delicate objects. In this paper, we design and fabricate a novel biomimetic tactile sensor with skin-like heterogeneity that perceives normal and shear contact forces simultaneously. It mimics the multilayers of mechanoreceptors by combining an extrinsic layer (piezoresistive sensors) and an intrinsic layer (a Hall sensor) so that it can perform estimation of contact force directions, locations, and joint-level torque. By integrating our sensors, a robotic gripper can obtain contact force feedback at fingertips; accordingly, robots can perform challenging tasks, such as tweezers usage, and egg grasping. This insightful sensor design can be customized and applied in different areas of robots and provide them with heterogeneous force sensing, potentially supporting robotics in acquiring skin-like tactile feedback.
\end{abstract}

\end{minipage}}}

\maketitle

\section{Introduction}
\label{sec:introduction}
\IEEEPARstart{H}{uman} skin can perceive essential somatosensory feedback via dense mechanoreceptors integrated in its multilayers and facilitate performing daily tasks such as grasping an egg and handing it over to a holder; nonetheless tasks like this still are challenging for contemporary robots, although robots already outperform humans in terms of vision, manipulation precision, and computational capacity \cite{billardTrendsChallengesRobot2019,goldsteinSensationPerception2016,shihElectronicSkinsMachine2020,johanssonCodingUseTactile2009,yangRobotLearningImitation2019,dahiyaTactileSensingHumans2010,yangElectronicSkinRecent2019,chortosSkininspiredElectronicDevices2014}. One of the main reasons behind this challenge is the lack of tactile sensing capabilities in robots that are equivalent to those of the human skin. Tactile sensing capabilities would be incredibly beneficial to robots in acquiring human skills by physically perceiving the contact forces (normal and shear forces), specifically those at the fingertips, when interacting with external environments \cite{shihElectronicSkinsMachine2020,nichollsSurveyRobotTactile1989,boutryHierarchicallyPatternedBioinspired2018,saudabayevSensorsRoboticHands2015}. In the human skin, four types of mechanoreceptors, namely Meissner’s corpuscles (fast-adapting, FA-I), Merkel receptors (slow-adapting, SA-I), Pacinian corpuscles (FA-II), and the Ruffini cylinder (SA-II) (illustrated in Fig. 1a), are highly integrated into hands (glabrous skin between the epidermis and dermis), which primarily respond to innocuous mechanical stimuli. The heterogeneous mechanoreceptors (FA-I and SA-II) in multilayer structures enable human hands to simultaneously perceive distributed normal force at the extrinsic layer (close to the epidermis) and gross shear force at the intrinsic layer (dermis); this is known as heterogeneity.

Recently, to improve the perception capabilities of robots, artificial tactile sensors have been widely developed via various sensing mechanisms, including piezoresistive \cite{sundaramLearningSignaturesHuman2019,leeNeuroinspiredArtificialPeripheral2019,zhangFingertipThreeaxisTactile2015}, capacitive \cite{boutryHierarchicallyPatternedBioinspired2018}, \cite{leporaTactileSuperresolutionBiomimetic2015,youArtificialMultimodalReceptors2020,ohScalableTactileSensor2020}, magnetic \cite{yanSoftMagneticSkin2021,tomoCoveringRobotFingertip2018,chathurangaMagneticMechanicalModeling2016}, barometric \cite{parkSoftnessadaptivePinchgraspStrategy2021,piacenzaDatadrivenSuperresolutionTactile2018}, and optical \cite{lambetaDIGITNovelDesign2020,yuanGelSightHighresolutionRobot2017,donlonGelSlimHighresolutionCompact,ward-cherrierTacTipFamilySoft2018} mechanisms. However, most conventional tactile sensors are developed based on a single sensing mechanism. Some of these sensors can only estimate normal force sensing and researchers have been improving such sensing capabilities by solely optimising post-processing procedures \cite{leporaTactileSuperresolutionBiomimetic2015,youArtificialMultimodalReceptors2020} or enhancing characteristics such as spatial resolution \cite{tomoCoveringRobotFingertip2018} and the force response curve (e.g., repetitivity, linearity, and sensitivity) \cite{ohScalableTactileSensor2020,yanSoftMagneticSkin2021,chathurangaMagneticMechanicalModeling2016,ward-cherrierTacTipFamilySoft2018}. Some studies leveraging a single sensing mechanism achieved normal and shear force sensing by decomposing force vectors from multiple distributed sensors around designated electrodes/microstructures \cite{boutryHierarchicallyPatternedBioinspired2018,zhangFingertipThreeaxisTactile2015,ohScalableTactileSensor2020}. Nonetheless, such approaches were limited by multiplied number of sensing elements when scaled up for real robotic applications. Recent popular optical-based tactile sensors \cite{donlonGelSlimHighresolutionCompact,ward-cherrierTacTipFamilySoft2018,wangGelSightWedgeMeasuring2021} can realize both distributed normal and shear contact force sensing with a spatial resolution comparable to that of human hands, but leaving enough space for light sources, pathway and receivers leads to their structural bulkiness that restricts their integration with anthropomorphic robotic hands, e.g., robotic prosthetic hands, having highly confined spacing. A soft skin \cite{tomoCoveringRobotFingertip2018} consisting of densely distributed 3-axis magnetic tactile sensors was developed to achieve large-area sensing for robotic hands with a compact size. However, these sensors would cause cross-coupling issues with adjacent elements. A promising research work \cite{yanSoftMagneticSkin2021} proposes a magnetic-based soft tactile sensor that can self-decouple normal and shear force sensing using a flexible sinusoidally magnetized film. However, its shear force sensing is limited to only a single direction due to its self-decouple magnetic film design, and its arrayed form unit that claims super-resolution (distributed force sensing) has cross-coupling issues with neighboring elements, and the actual spatial resolution of its single unit is limited.

To tackle these challenges, we developed a novel biomimetic heterogeneous tactile sensor, GTac, that can achieve skin-like heterogeneity to perceive distributed normal and gross shear contact forces simultaneously. The design of GTac maintains high integrability, mimicking the multilayers of the mechanoreceptors (FA-I and SA-II) between the epidermis and dermis of human skin by using a similar arrangement comprising an extrinsic layer (arrayed piezoresistive sensors) and an intrinsic layer (a Hall sensor).

\section{Methods}
\subsection{Study design}
Our objectives were to show that GTac can achieve human-like heterogeneous tactile sensing capabilities based on the skin-like multilayer structure maintaining integrability. In the characterization experiments, we collected data for the FA-I and SA-II layers to evaluate the contact force estimation performance of GTac. In our experiments, the magnet in the SA-II layer of GTac was chosen according to the magnetic disturbance analysis. To verify the integrability of GTac, we integrated customized GTac into the fingertips of a gripper for executing object grasping tasks.

\subsection{Design of the bio-inspired tactile sensor}
\begin{figure}[!t]
\centerline{\includegraphics[width=\columnwidth]{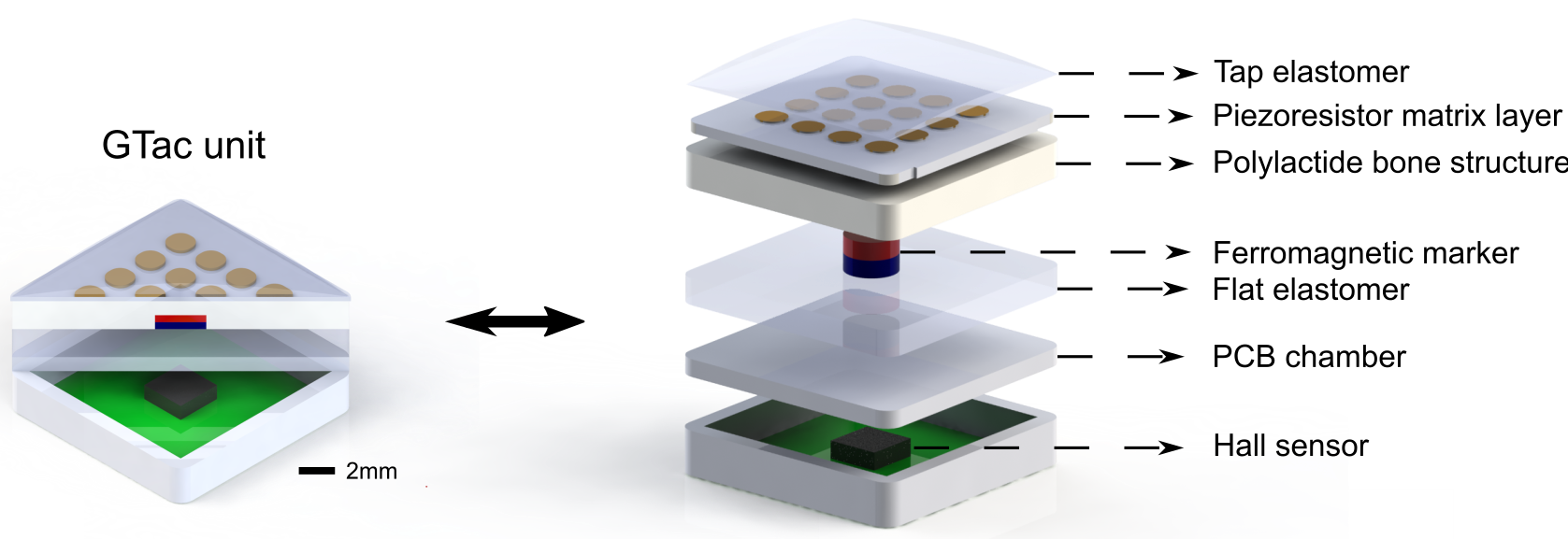}}
\caption{Multilayer structure and characterization performance of GTac. Left: Section view of the skin-inspired multilayer structure model; Right: Components of a GTac unit.}
\label{fig_GTac_design}
\end{figure}

Inspired by the tactile sensing mechanism and multilayer structure of human skin, we developed a biomimetic tactile sensor, GTac (Fig. \ref{fig_GTac_design}). GTac can estimate the extrinsic arrayed normal contact force in its FA-I layer (0.5 mm thick, the epidermis of the human hand is ~1.5 mm thick), the tri-axis gross contact force in its SA-II layer (3 mm thick, the dermis of human skin is ~1–4 mm thick) \cite{goldsteinSensationPerception2016}, and the joint-level torque in both the FA-I and SA-II layers via force estimation, based on known geometrical dimensions when an external contact force is applied on the top layer. More specifically, GTac transforms the normal extrinsic contact force into the reduction of the resistance and transforms the gross tri-axis force into a local magnetic flux density change. The resistance and local magnetic flux density can be measured using piezoresistive sensors and Hall sensors independently and simultaneously. The circuits used for producing the analog and digital signals were embedded in the printed circuit board (PCB) at the bottom layer.

\subsection{Heterogeneous force feedback}
Humans can approximate the location of the center of the contact force on their fingertips \cite{johanssonCodingUseTactile2009}. Accordingly, we also explored the contact location estimation capability of GTac by performing a weighted average of the detected pressure at each sensing point.
\begin{equation}\left[ \begin{array}{l}
{x_e}\\
{y_e}
\end{array} \right] = \left[ \begin{array}{l}
{{\sum\nolimits_{r = 1}^4 {\sum\nolimits_{c = 1}^4 {(e \cdot c \cdot {R^{r,c}})} } } \mathord{\left/
 {\vphantom {{\sum\nolimits_{r = 1}^4 {\sum\nolimits_{c = 1}^4 {(e \cdot c \cdot {R^{r,c}})} } } {16}}} \right.
 \kern-\nulldelimiterspace} {16}}\\
{{\sum\nolimits_{r = 1}^4 {\sum\nolimits_{c = 1}^4 {(e \cdot r \cdot {R^{r,c}})} } } \mathord{\left/
 {\vphantom {{\sum\nolimits_{r = 1}^4 {\sum\nolimits_{c = 1}^4 {(e \cdot r \cdot {R^{r,c}})} } } {16}}} \right.
 \kern-\nulldelimiterspace} {16}}
\end{array} \right],\label{loc_est}
\end{equation}
where ${R^{r,c}}$ is the signal reading from the FA-I layer in row $r$ and column $c$, and $e$ is the spatial resolution of the FA-I layer, $e = 2.5mm$. Like human cutaneous softness, the external contact force can deform the elastomer substrate on the contact surface. This elastomer mechanically buffers the pressure in the case of sharp contact and thereafter delivers the pressure from the contact surface to the FA-I layer whose electrical resistance is reduced because of the mechanical strain. Moreover, the contact force is transmitted to the bone structure, deforming the flat elastomer relative to the base chamber. Therefore, the bone structure can move along the tri-axis relative to the Hall effect sensor, changing the local magnetic flux density. This change in the local magnetic flux density can be measured by the Hall sensor and used to estimate the shear contact force. The representation of the composition of force sensing on a GTac is the hybrid result of the FA-I and SA-II signals. The linear relationship between the tri-axis contact force and sensing signals from GTac can be expressed as

\begin{equation}
\resizebox{0.9\hsize}{!}{$\left[ \begin{array}{l}
{F_x}\\
{F_y}\\
{F_z}
\end{array} \right] = \left[ \begin{array}{l}
k{}_x\Delta {B^x} + {b_x}\\
k{}_y\Delta {B^y} + {b_y}\\
k{}_z(a\Delta {B^z} + (1 - a)\sum\limits_{r = 1}^4 {\sum\limits_{c = 1}^4 {{R^{r,c}}} } ) + {b_z}
\end{array} \right]$}
,
\label{force_est}
\end{equation}
where $\Delta {B^{x,y,z}}$ is the observed change in the local magnetic flux density relative to the initialized position, and ${k_{x,y,z}}$ and ${b_{x,y,z}}$ are the slope and intercept of the linear fitting lines in tri-axis, respectively. Since the FA-I and SA-II layers have a redundant force sensing degree of freedom (DoF) in the z-axis, we parameterized both force sensing sources in the z-axis, i.e., $a\Delta {B^z} + \left( {1 - a} \right)\mathop \sum \nolimits_{r = 1}^4 \mathop \sum \nolimits_{c = 1}^4 {R^{r,c}}$ in Equation (2), where $\mathop \sum \nolimits_{r = 1}^4 \mathop \sum \nolimits_{c = 1}^4 {R^{r,c}}$ is the sum of arrayed FA-I signals in the $4 \times 4$ matrix, to address the redundancy.

\subsection{Principle}

\begin{figure}
\centerline{\includegraphics[width=\columnwidth]{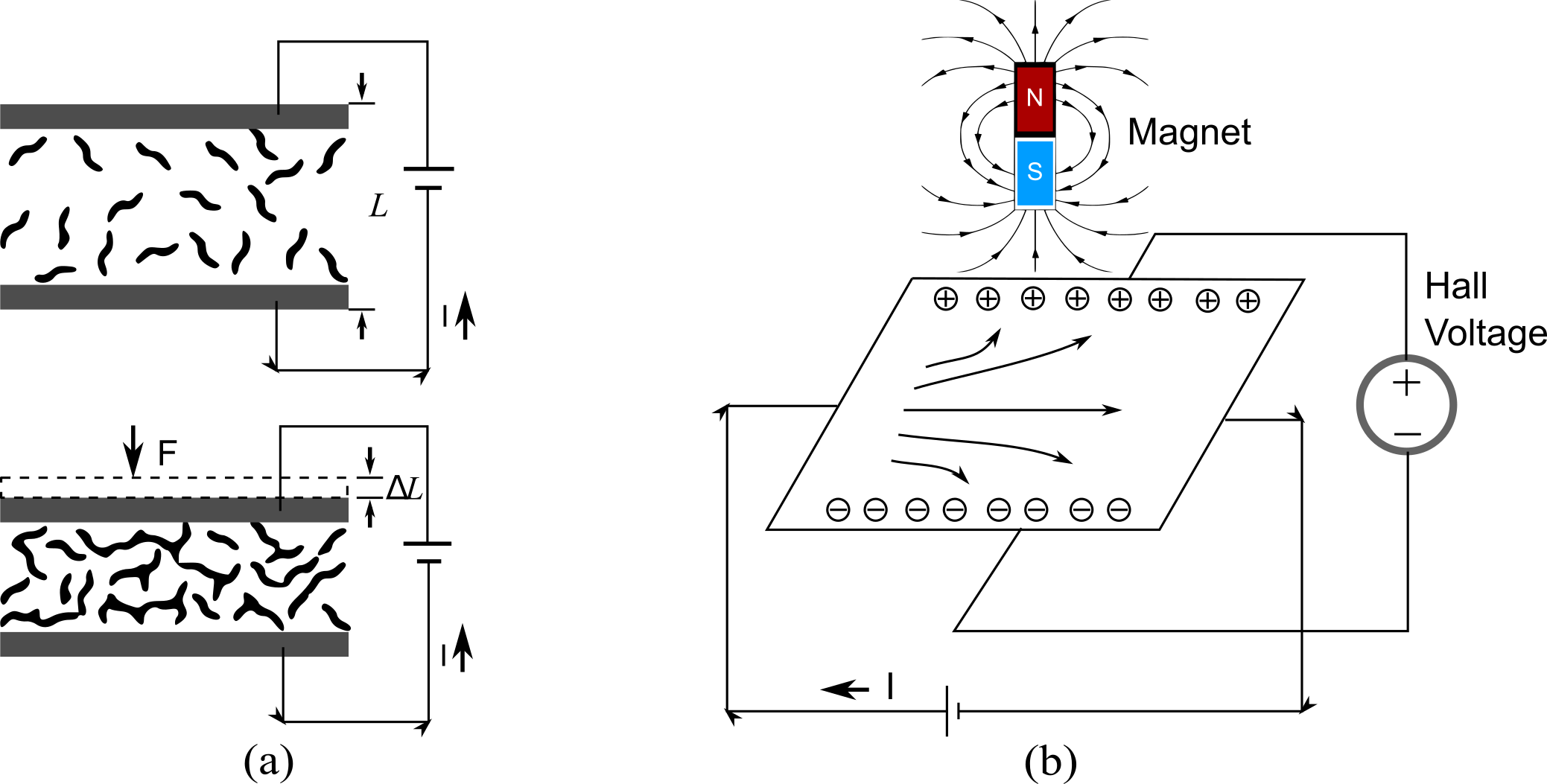}}
\caption{(a) Illustration of piezoresistive effect, where a mechanical stress is applied on the piezoresistors, and the electrical resistivity decreases due to the deformed inter-atomic spacing of the conducting materials. (b) Illustration of Hall effect, wherein Hall voltage is produced when the charge carriers with movement altered by the Lorentz force from the present magnetic flux, are in contact with the edges of the void.}
\label{fig_principles_circuits}
\end{figure}

Piezoresistive effect: The piezoresistive effect explains that the resistance of piezoresistors is reduced owing to the decreased interatomic spacing of conducting materials when a mechanical stress is applied (Fig. \ref{fig_principles_circuits}(a)). 

\begin{equation}
\left\{ \begin{array}{l}
\varepsilon  = \frac{{\Delta L}}{L}\\
\frac{{\Delta R}}{R} = K \cdot \varepsilon 
\end{array} \right],
\label{piezo_principle}
\end{equation}
where $K$ is a constant indicating the Gauge factor of the piezoresistors. $\varepsilon$ is the mechanical strain applied. $L$ is the original thickness and $\Delta L$ is the absolute change in thickness. The electrical resistance of the sensors is proportional to the strain applied to the sensors that can be obtained using a follower-based signal isolation circuit (Fig. \ref{fig_FAI_circuits}).

Hall effect: the Hall effect refers to a voltage difference, called the Hall voltage, that is produced when the charge carriers with movement are altered by the Lorentz force from the present magnetic flux flow to the edges of the void (Fig. \ref{fig_principles_circuits}). $V = {IBl}/{neA}$ derives the Hall Voltage V that is proportional to the local magnetic density B. In the Hall effect sensor, I, l, n, e, and A are the internal configurations representing the current in the sensing strip, the length of the sensing strip, the number of charge carriers per unit volume, the charge amount per charge carrier, and the area of the sensing strip of the Hall sensor, respectively. Here, we can observe the local change in the magnetic flux density by the Hall sensor for estimating the contact force via the digital output from the Hall sensor, i.e., MLX90393, as the SA-II layer signals.

\begin{figure}[!h]
\centerline{\includegraphics[width=\columnwidth]{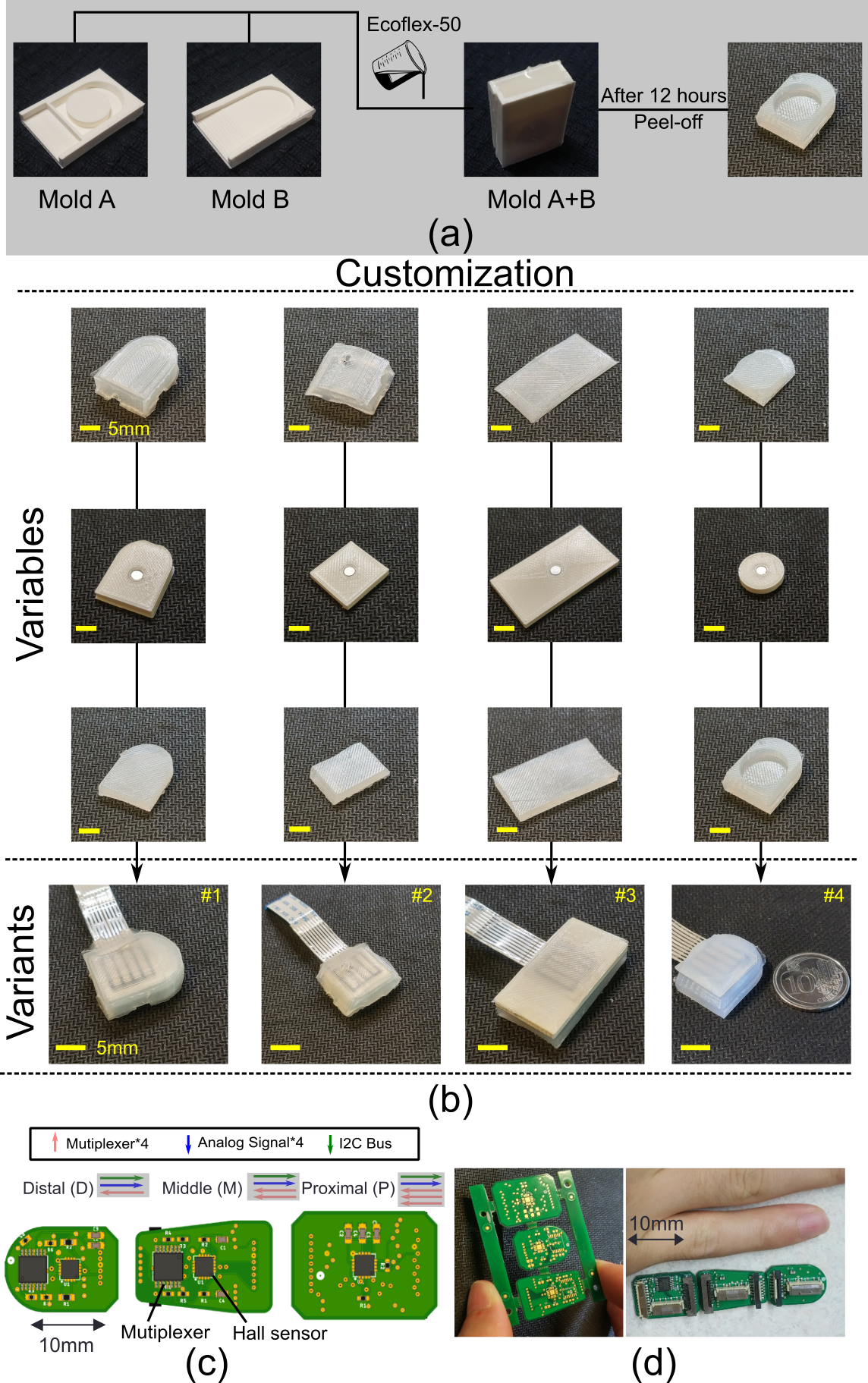}}
\caption{(a) The process of making a silicone rubber part of GTac using 3D printed molds. (b) Four various customized GTac. (c) The 3 variant GTac in each set and their serial signals transimission prototype. (d) The manufactured PCB of GTac.}
\label{fig_fabrication_custm}
\end{figure}

\subsection{Fabrication}
The design of the biomimetic tactile sensor, GTac, adopts a multilayer structure. As shown in Fig. \ref{fig_fabrication_custm}, each GTac unit consists of two elastomer substrates, a piezoresistive sensor laminate (4 × 4), a polylactide bone structure, a cylinder neodymium (NdFeB) magnet, a base chamber, and a PCB with an integrated Hall sensor. The height and diameter of the cylinder magnet used in this study were 1 and 3 mm, respectively. The arrayed normal force sensing laminate, i.e., piezoresistive sensors, forms the first layer, and an elastomer substrate is placed on one side of it as the soft contact surface. The other side of the laminate is attached to the next layer, a rigid bone structure magnetized by mounting the cylindrical NdFeB magnet. Under this layer, there is a flat elastomer substrate that is attached to the base chamber that integrates the Hall effect sensor. All the elastomer substrates were made by casting and drying (12 hours) liquid silicone rubber (Ecoflex 00-50, Smooth-On Inc.) in customized molds (Fig. \ref{fig_fabrication_custm}(a)), with an elastic modulus of 83 kPa. All the layers were stuck together using an adhesive, Cyanolube Cyanoacrylate (Electrolube Co. Ltd). The polylactide bone structures and customized molds were printed by the Ultimaker S5 using PLA White (Ultimaker Co. Ltd.). Arrayed piezoresistive sensors (Roxifsr Co. Ltd.) and cylinder magnets (PinChun Co. Ltd.) were readily available. The Hall sensor was an MLX90393 integrated circuit chip (Melexis Co. Ltd.). The MLX90393 Qwiic Magnetometer (SparkFun Co. Ltd.) was used in the characterization experiments. To fabricate the structures of the two-fingered gripper, we used the Ultimaker S5 printer (material: White PLA).

\subsection{Magnetic disturbances cancellation}
As the sensing principle of the SA-II layer is based on magnetic flux density measurement, there are magnetic disturbances ($d$) from two main sources, the earth’s magnetic field and adjacent magnetic field. To quantify the magnetic disturbance, we term an alternative definition of signal-to-noise-ratio (SNR) as $SNR = s/(s + d)$, where $s$ is the effective signal strength. The earth’s magnetic field is a constant vector (${{\bf{B}}_{\bf{e}}}$) in the environment, but unknown in the beginning (we can observe its magnetic flux density change via Hall sensor); thus, it is included in the sensor observation (${{\bf{B}}_{\bf{s}}}$), i.e., ${{\bf{B}}_{\bf{s}}}{\bf{ = }}{{\bf{B}}_{\bf{e}}}{\bf{ + }}{{\bf{B}}_{\bf{m}}}$, where ${{\bf{B}}_{\bf{m}}}$ is the magnetic field of the magnet in GTac. Hence, ${{\bf{B}}_{\bf{m}}}{\bf{ = }}{{\bf{B}}_{\bf{s}}}{\bf{ - }}{{\bf{B}}_{\bf{e}}}$, and the contact force estimation is only related to ${{\bf{B}}_{\bf{m}}}$, i.e., ${\bf{F}} = f\left( {{{\bf{B}}_{\bf{m}}}} \right) = f\left( {{{\bf{B}}_{\bf{s}}}{\bf{ - }}{{\bf{B}}_{\bf{e}}}} \right)$. Therefore, the solution is to determine ${{\bf{B}}_{\bf{e}}}$ and subtract it for the contact force estimation. Using the matrix multiplication of the rotation matrix, a contact vector ${q_b}$ in the new coordinate after orientation ${{\bf{R}}_{{\bf{ab}}}}$ can be obtained using ${{\bf{q}}_{\bf{a}}}{\bf{ = }}{{\bf{R}}_{{\bf{ab}}}}{{\bf{q}}_{\bf{b}}}$. Accordingly, we obtain${\bf{\Delta q = }}{{\bf{q}}_{\bf{a}}}{\bf{ - }}{{\bf{q}}_{\bf{b}}}{\bf{ = }}\left( {{{\bf{R}}_{{\bf{ab}}}}{\bf{ - I}}} \right){{\bf{q}}_{\bf{b}}}$. Therefore, the constant vector of the earth’s magnetic field in the environment ${{\bf{B}}_{{\bf{e|b}}}}$ can be obtained by solving the linear equation via least-squares optimization:
\begin{equation}
    	\left\{ \begin{array}{l}
\Delta {{\bf{B}}_{\bf{s}}} = ({{\bf{R}}_{{\bf{ab}}}}{\bf{ - I}}){{\bf{B}}_{{\bf{e|b}}}}\\
{{\bf{B}}_{{\bf{e|b}}}} = {[{({{\bf{R}}_{{\bf{ab}}}}{\bf{ - I}})^T}({{\bf{R}}_{{\bf{ab}}}}{\bf{ - I}})]^{ - 1}}{({{\bf{R}}_{{\bf{ab}}}}{\bf{ - I}})^T}\Delta {{\bf{B}}_{\bf{s}}}
\end{array} \right],
\label{disturb_cancle}
\end{equation}

where ${{\bf{R}}_{{\bf{ab}}}}$ is the rotation matrix and can be obtained via inertial measurement unit (IMU) or angle encoders, and ${\bf{\Delta }}{{\bf{B}}_{\bf{s}}}$ is the observed magnetic flux density change by the sensor. 

\section{Experiments}
\subsection{Sensor characterization}

\begin{figure}
\centerline{\includegraphics[width=\columnwidth]{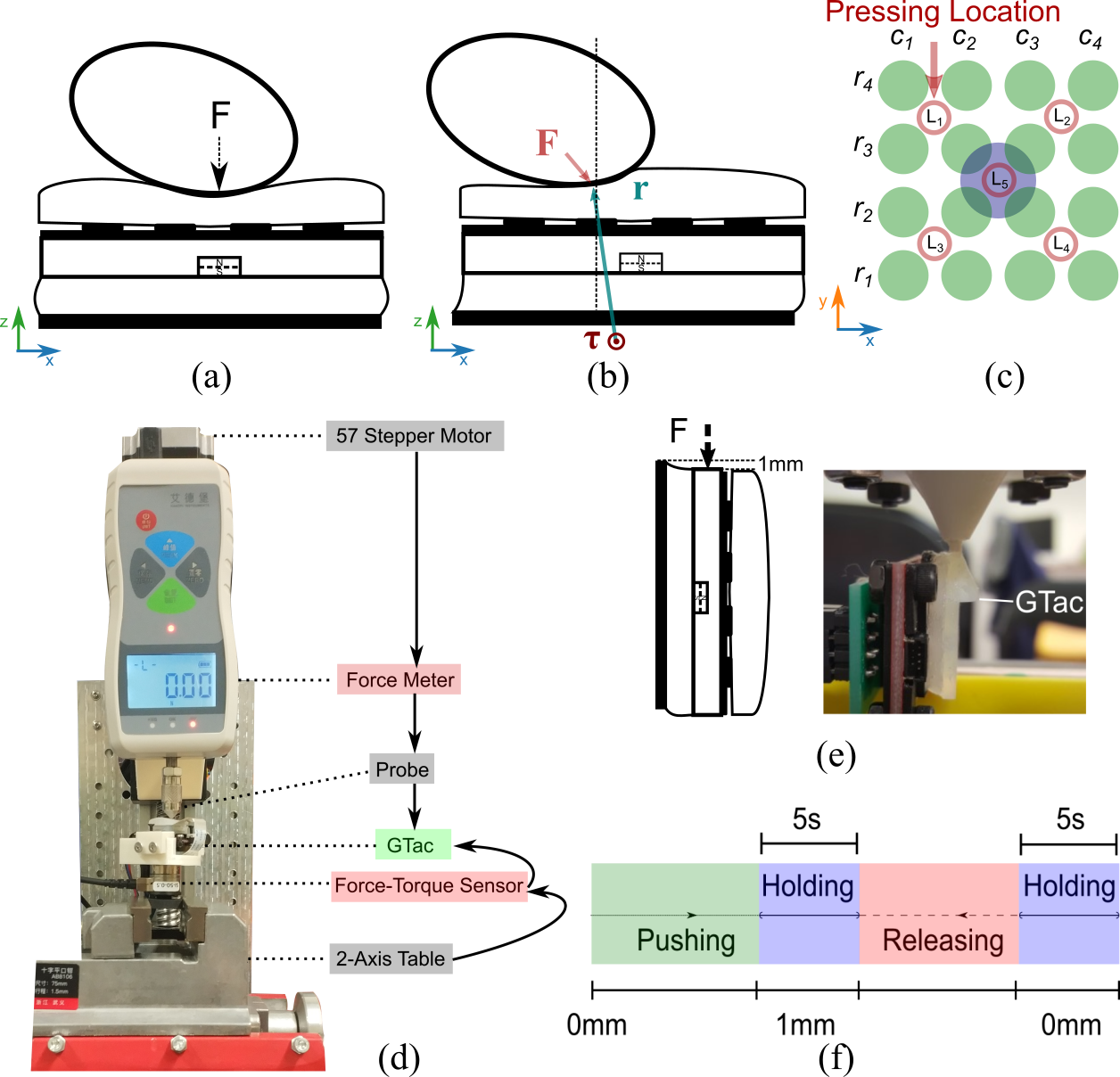}}
\caption{(a)-(b) Illustration of the pressing force (normal and shear forces) applied on GTac which deforms the silicone rubber parts. (c) The probe applies pressing force the during the characteristics experiments at the five distinct contact locations, i.e., L1–L5. (d) Three-axis data acquisition platform for GTac characterization. (e) Experimental setup for collecting shear force data from SA-II layer. (f) The periodic pressing and releasing cycling during the data collection for SA-II layer characterisation.}
\label{fig_characteristic_setup}
\end{figure}

A data collection platform (Fig. \ref{fig_characteristic_setup}) was built to quantitatively measure the contact force applied to the GTac sensor. This platform comprised a force meter (Handpi, Co. Ltd) to measure the normal extrinsic force, a 6-axis force torque (F-T) sensor (Nano17, ATI Co. Ltd.), a 2-axis manual motion table, and a ball screw that can move the force meter vertically (precision: 0.05 mm). A cone-shaped probe (height: 3.8 mm; radius: 5.3 mm) was used to vertically press the sensor at five primary distinct locations (L1–L5) to collect data. The contact force was incremented by 0.25 N from 0 to 2 N according to the force meter. The F-T sensor was mounted vertically 10 mm below the center of the FA-I layer of GTac and it was fixed by a 2-axis motion table that was used to manually switch the pressing location.

\subsection{Data acquisition and processing}
\begin{figure}
\centerline{\includegraphics[width=\columnwidth]{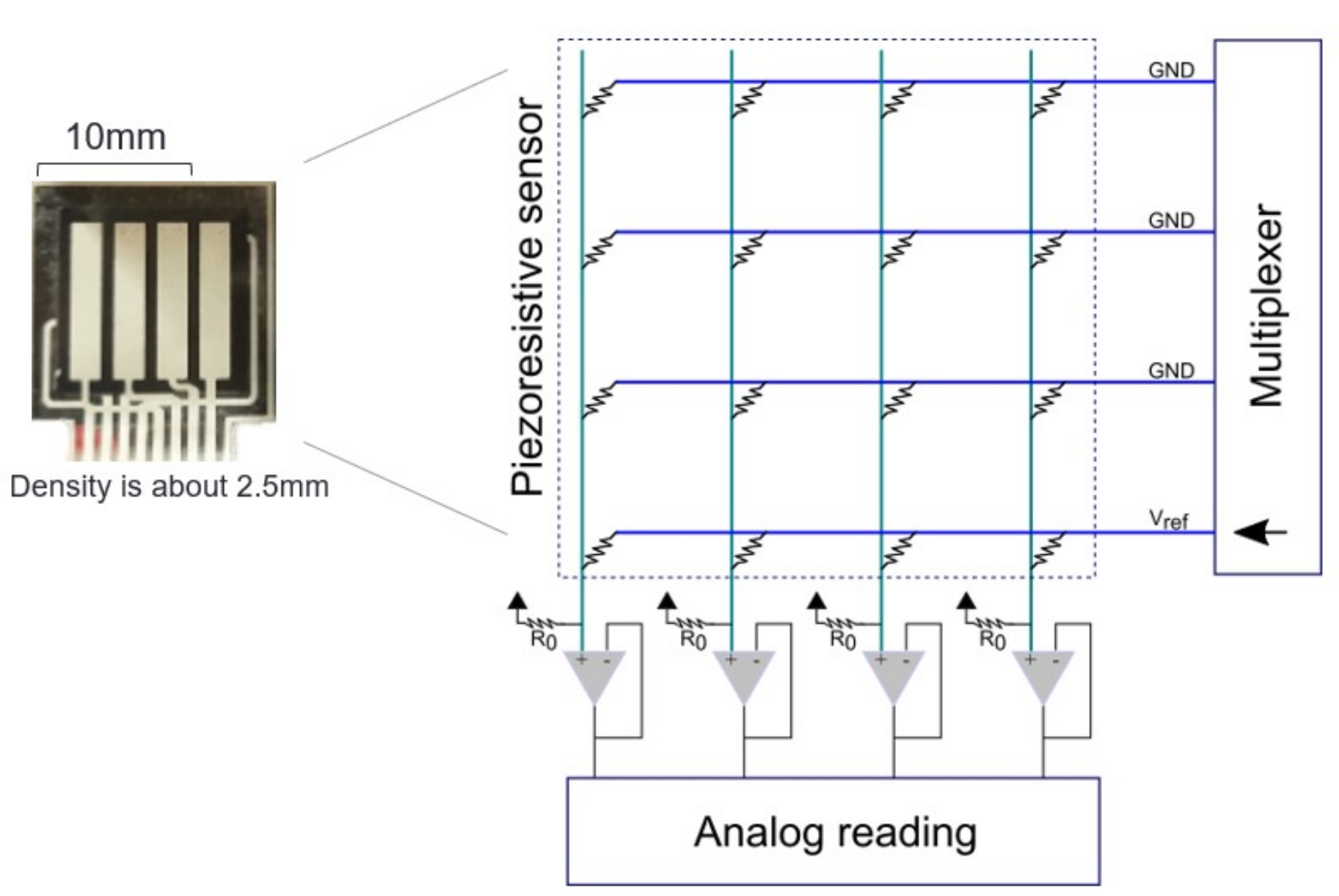}}
\caption{Left: The photograph of 4 by 4 matrix piezoresistive sensors as the FA-I layer in GTac whose spatial resolution/density is about 2.5mm; Right: Illustration of the circuit schematic, in one finger section, to acquire the analog output of arrayed piezoresistive sensors utilizing the follower-based signal isolation circuit. Only one row would be connected to ${V_{ref}}$ at each moment and the other would be short to GND.}
\label{fig_FAI_circuits}
\end{figure}

The gripper with 2 integrated GTac could acquire 38 tactile signals at 250 Hz. Initialization: according to the sensing principles of GTac, the change in the local magnetic flux density was measured by the Hall sensor. Since the Hall sensor measured the global magnetic flux density, we   initialized GTac by collecting ${N_0}$ samples (we set ${N_0}$ as 300) and storing the mean value of the tail 100 samples (0.67 second data). Subsequently, all data were subtracted from the mean values obtained in the initialization to obtain the relative values of the tactile sensing signals. To reduce the noise of the signals, a moving average was added to filter all the signals with a window size of 6. As shown in Fig. \ref{fig_FAI_circuits}, the circuit schematic in each finger section acquires the analog output of 4 × 4 matrix piezoresistive sensors, utilizing the follower-based signal isolation circuit. Only one row is connected to ${V_{ref}}$  at each moment, and the rest is short to ground (GND).

\subsection{Magnetic disturbance observation}
\begin{figure}[h]
\centerline{\includegraphics[width=\columnwidth]{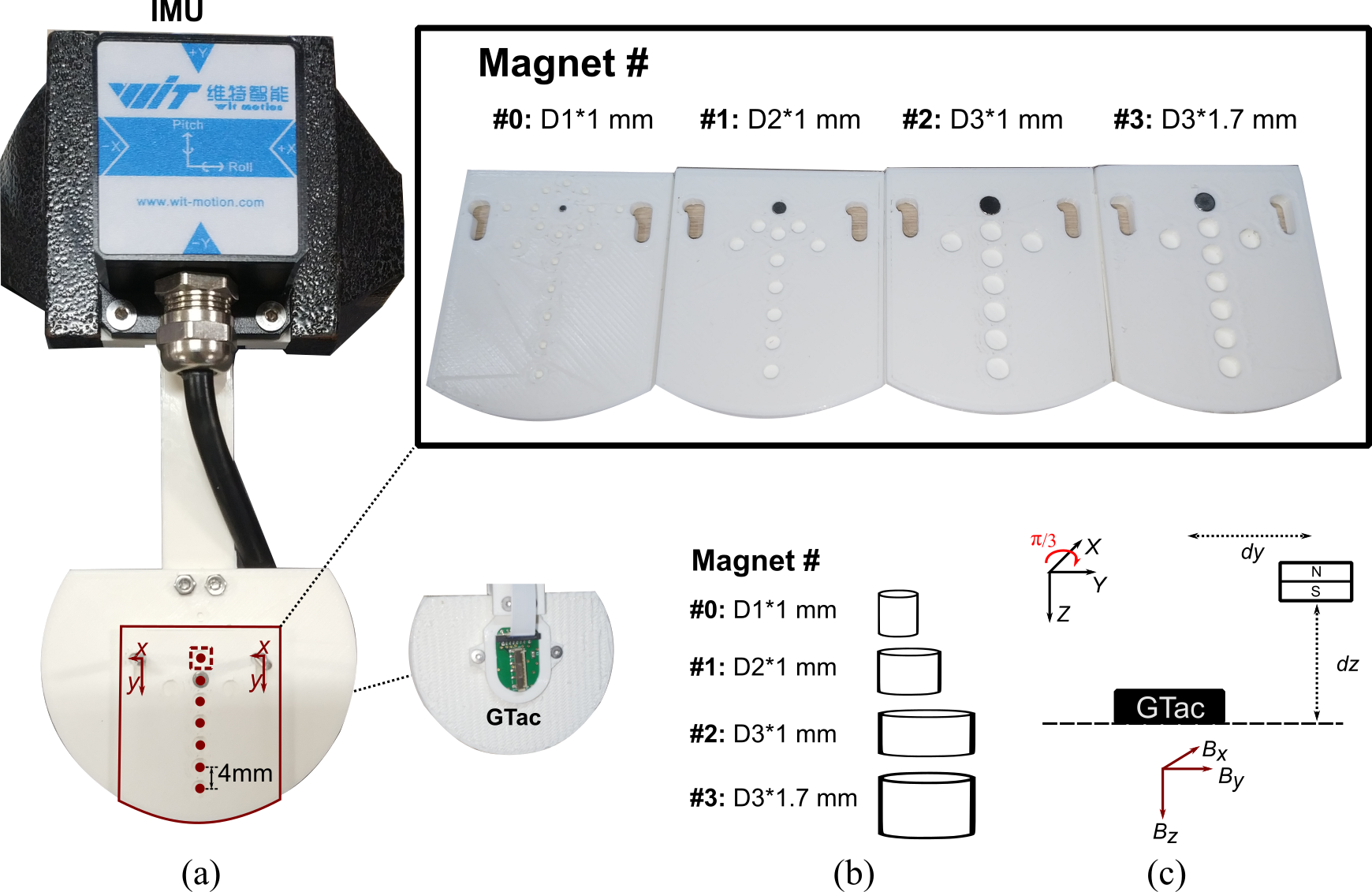}}
\caption{(a) Experimental setup consists of a GTac, an IMU, and the frame with a specific gap (4mm) for placing the magnets. All the parts are mounted together and used to collect the SA-II data from the GTac and IMU. (b) Four different candidate magnetic markers of different sizes. (c) Illustration of coordinate used in the experiment.}
\label{fig_disturb_setup}
\end{figure}

To observe the magnetic disturbances on the SA-II type signals of GTac, we built an experimental setup consisting of an IMU, a GTac unit, four magnets with various sizes, and holed bases with certain gaps (Fig. \ref{fig_disturb_setup}). The holed bases can contain the magnets and move the magnets by 1 mm and 4 mm in x-axis and y-axis respectively. The effective signal strengths were recorded while the magnets were moved by 1 mm in x-axis and the simulated adjacent disturbances were recorded while the magnets were displaced by 4 mm in y-axis from the pre-set starting points. The disturbances from earth’s field were recorded while the GTac was rotated by $\pi /3$ along its x-axis. Therefore, the SNR under the two types of magnetic disturbances were calculated.

\subsection{Control strategy}
To achieve closed-loop grasping using the gripper with integrated GTac to grasp fragile objects, a threshold ${T_g}$ was used to control the gripping force exerted by the fingers via feedback from GTac on each fingertip. To grasp the object, the corresponding motor, ${m_f}$ of each finger $f$, rotated 1 increment (1.5º) to drive the rack and pinion gear to conduct finger closure until the leveraged GTac signals ${g_f} > {T_g}$. The leveraged GTac signals can be derived by 
\begin{equation}
\resizebox{.9\hsize}{!}{${g_f} = \sqrt {\Delta B{{_f^x}^2} + \Delta B{{_f^y}^2} + {{[a\Delta B_f^z + (1 - a)\sum\limits_{r = 1}^4 {\sum\limits_{c = 1}^4 {R_f^{r,c}} } ]}^2}}$},
\label{gf_cal}
\end{equation}
where $f = 1$ denotes the left finger and $f = 2$ denotes the right finger. In tweezers usage experiment, we set ${T_g} = 700$ and $a = 0.3$ for egg grasping experiments. We set ${T_{g_{high}}} = 900$, ${T_{g_{low}}} = 500$ and $a = 0.3$ for tweezers grasping experiments. When $g_1<T_{g_{high}}$ and $g_2<T_{g_{high}}$, both fingers started closing until $g_1>T_{g_{high}}$ and $g_2>T_{g_{high}}$. After 2 seconds, both fingers started releasing the object but holding the tweezers until $g_1<T_{g_{low}}$ and $g_2<T_{g_{low}}$. The robotic arm (UR10) used in the experiments was moving in a pre-programmed trajectory.

\section{RESULTS}
\subsection{Estimation of contact forces}

\begin{figure}
\centerline{\includegraphics[width=0.5\columnwidth]{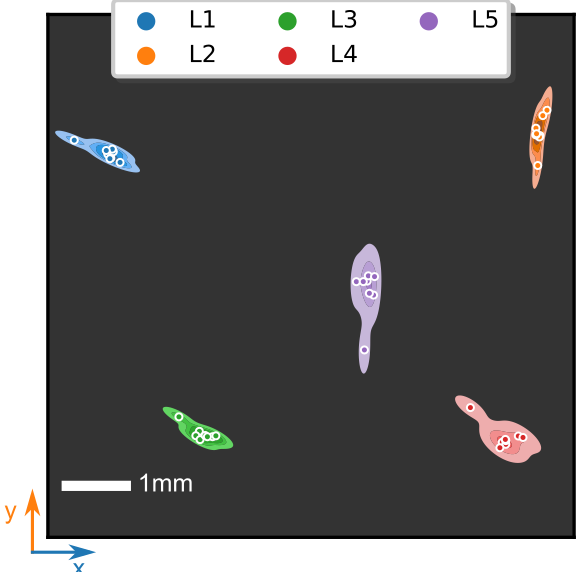}}
\caption{Contact locations estimated by GTac from L1 to L5.}
\label{fig_loc_est}
\end{figure}

\begin{figure*}[!ht]
\centerline{\includegraphics[width=\textwidth]{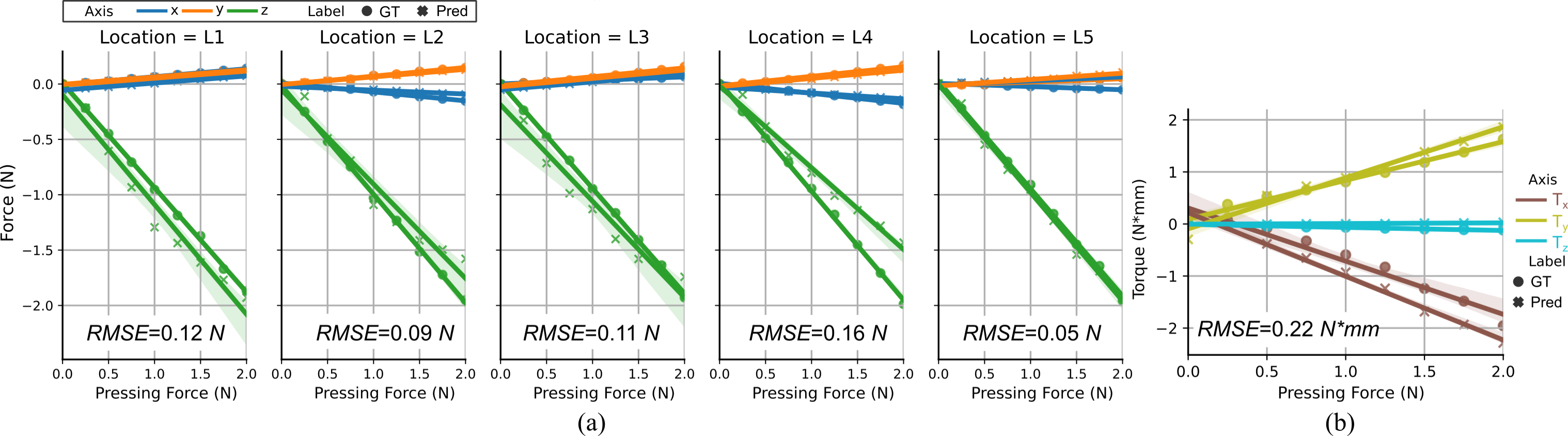}}
\caption{(a) Contact force estimated by GTac relative to the ground truth in tri-axis from L1 to L5.  (b) Torque estimation results from using GTac and comparison with the ground truth. All the line shadows indicate a confidence interval.}
\label{fig_force_torq_est}
\end{figure*}

\begin{figure}
\centerline{\includegraphics[width=\columnwidth]{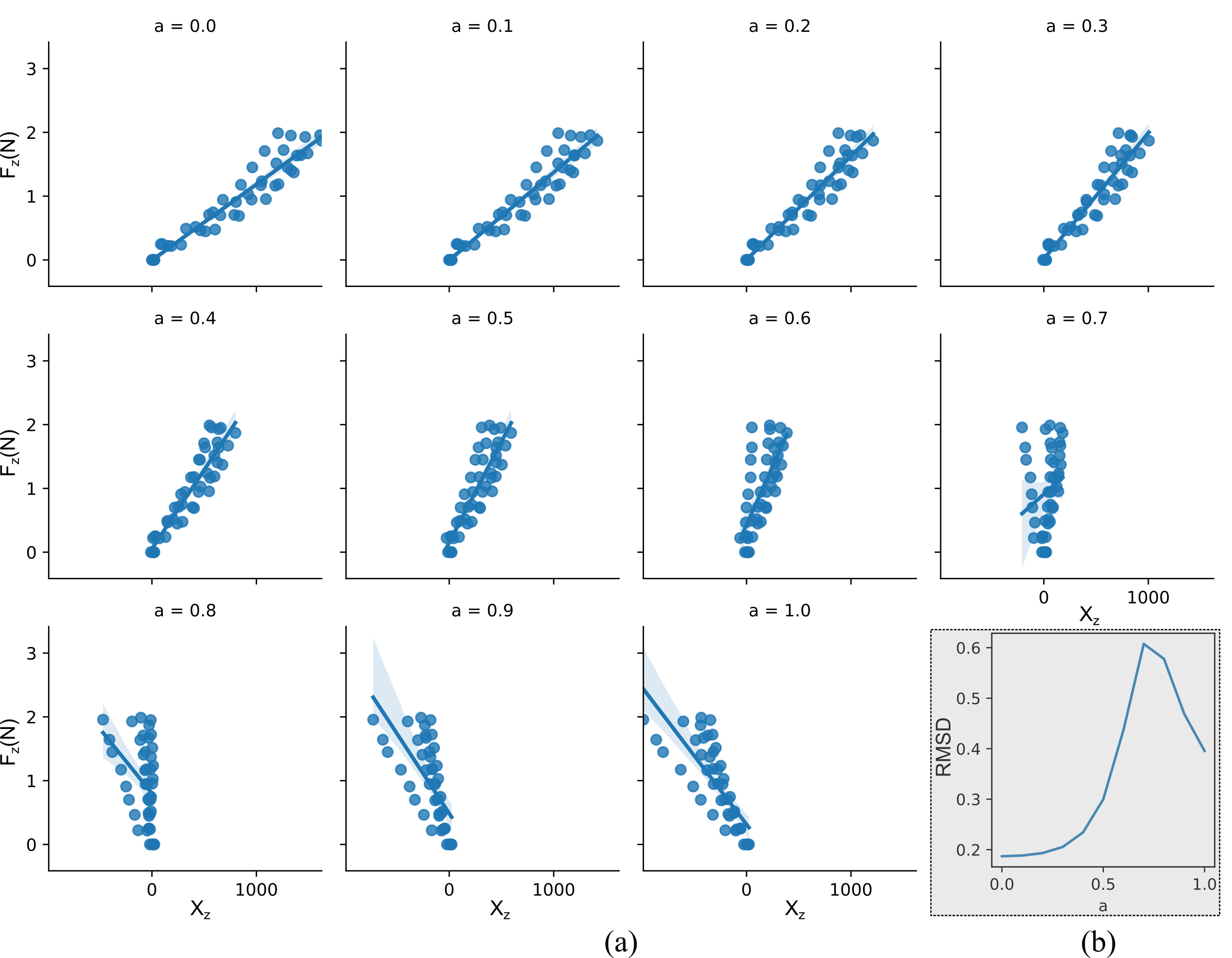}}
\caption{(a) Linearity analysis of normal force estimation. a Fitting line of ground truth, ${F_z}$, and the parameterized force estimation in z-axis, ${X_z} = a\Delta {B^z} + \left( {1 - a} \right)\mathop \sum \nolimits_{r = 1}^4 \mathop \sum \nolimits_{c = 1}^4 {R^{r,c}}$ at various a. (b) Root-mean-square deviation (RMSD) of the force estimation in z-axis, where the RMSD is minimum at  $a = 0$.}
\label{fig_Fz_a}
\end{figure}

\begin{figure}
\centerline{\includegraphics[width=\columnwidth]{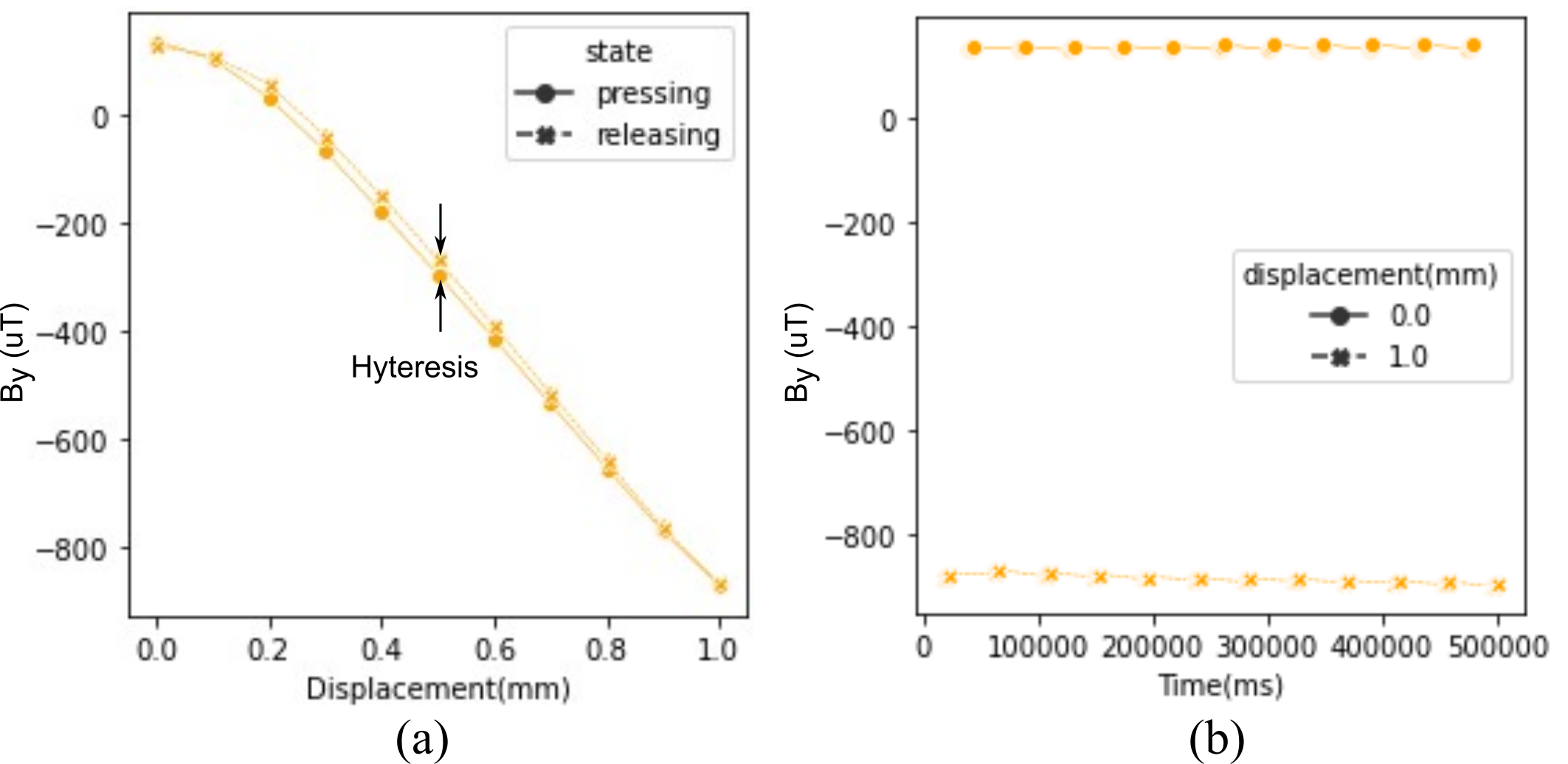}}
\caption{(a) Curve of detected local magnetic flux density versus displacement in the pressing direction. (b) Repetitivity results of the SA-II layer of GTac.}
\label{fig_hysteresis}
\end{figure}

Regarding the coupling issue in magnetic sensors \cite{yanSoftMagneticSkin2021}, GTac can naturally decouple the normal and tangential force sensing using a regular cylindrical magnetic marker because the FA-I and SA-II layers of GTac perceive tactile information independently. The root mean square error (RMSE) of the contact location estimation was [0.88, 0.90, 0.98, 0.38, 0.88] \textit{mm} at locations L1-L5 (Fig. \ref{fig_loc_est}) according to the data we collected from GTac and the force-torque sensor. The RMSE of the tri-axis force estimation was [0.12, 0.09, 0.11, 0.16, 0.05] \textit{N} at locations L1–L5 (Fig. \ref{fig_force_torq_est}(a)), where we set ${\rm{a}} = 0$ according to the linearity analysis (Fig. \ref{fig_Fz_a}). As shown the hysteresis and repetitivity test in Fig. \ref{fig_hysteresis}, the maximum hysteresis error is 2\% and the non-linearity occurs at the beginning 0.2 \textit{mm}. Since the contact location and tri-axis force ${\bf{F}}$ could be estimated by GTac, assuming there is a virtual joint center to which GTac is connected, then GTac could estimate the joint-level torque, ${\bf{\tau }} = {\bf{r}} \times {\bf{F}}$, exerted by the external force (Fig. \ref{fig_characteristic_setup}(b)), where ${\bf{r}}$ is the force arm from the estimated contact location to the virtual joint center. As the results show (Fig. \ref{fig_force_torq_est}(b)), the RMSE of the torque estimation is $0.22N \cdot mm$ in the tri-axis.

\subsection{Magnetic disturbance cancellation}
\begin{figure}
\centerline{\includegraphics[width=\columnwidth]{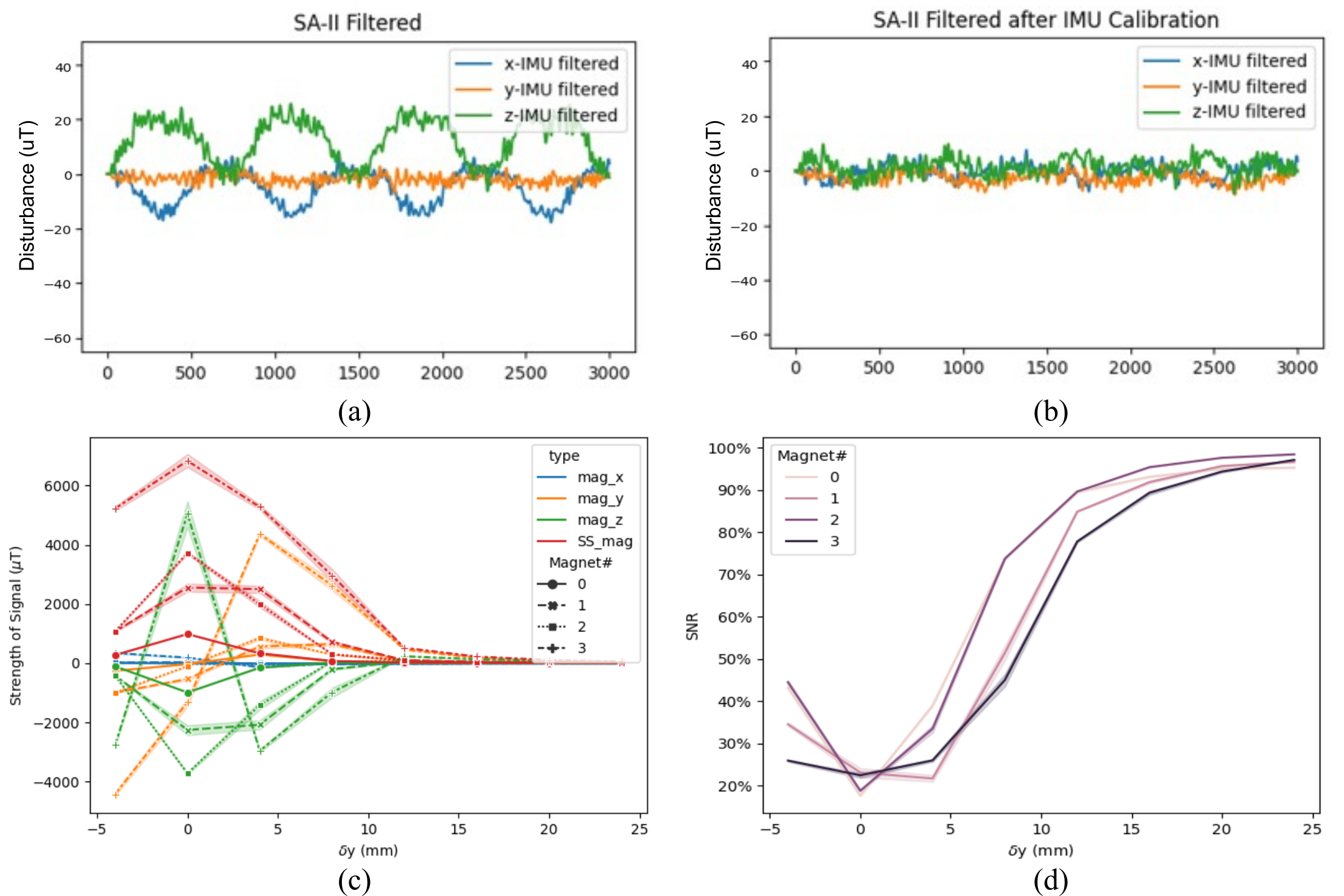}}
\caption{(a) Observed disturbance from the earth’s magnetic field when periodically rotating a GTac unit by $\pi /3$. (b) Observed disturbance filtered by IMU calibration method. (c) Relationship of the detected magnetic flux density with displacement along y-axis for four different magnets. (d) Relationship of SNR versus displacement along y-axis with the measured effective signals.}
\label{fig_disturb_results}
\end{figure}

We recorded the disturbance (filtered by moving average of window size 6) when rotating the GTac repeatedly $\pi /3$ along its x-axis (Fig. \ref{fig_disturb_results}(a)), which reduces the SNR by 5.3\%. The magnetic disturbance from the adjacent magnet is due to the local magnetic field detected by the adjacent GTac units integrated into robots, e.g., robotic hand. Therefore, the disturbance strength is related to the magnetic strength of the magnets used in the GTac. We estimate the relationship of the detected magnetic flux density with the displacement along y-axis, $\delta y$, using four magnets with different magnetic strengths, and the magnetic flux density change of 1-mm displacement along x-axis is regarded as its effective signal $s$. The measured effective signals of four magnets, i.e., $[211\mu T,580\mu T,853\mu T,1816\mu T]$. Magnet \#2 was found to have the highest SNR according to the measured relationship of the SNR versus $\delta y$ (Fig. \ref{fig_disturb_results}(d)) when the displacement $\delta y \ge 16mm$. The SNR can be maintained above 93\% from adjacent disturbances. Our solution utilizing IMU calibration can reduce the disturbance (Fig. \ref{fig_disturb_results}(b)), which improve the SNR by 69.8\% of the reduced SNR by disturbance from earth’s field.

\begin{figure*}[!ht]
\centerline{\includegraphics[width=\textwidth]{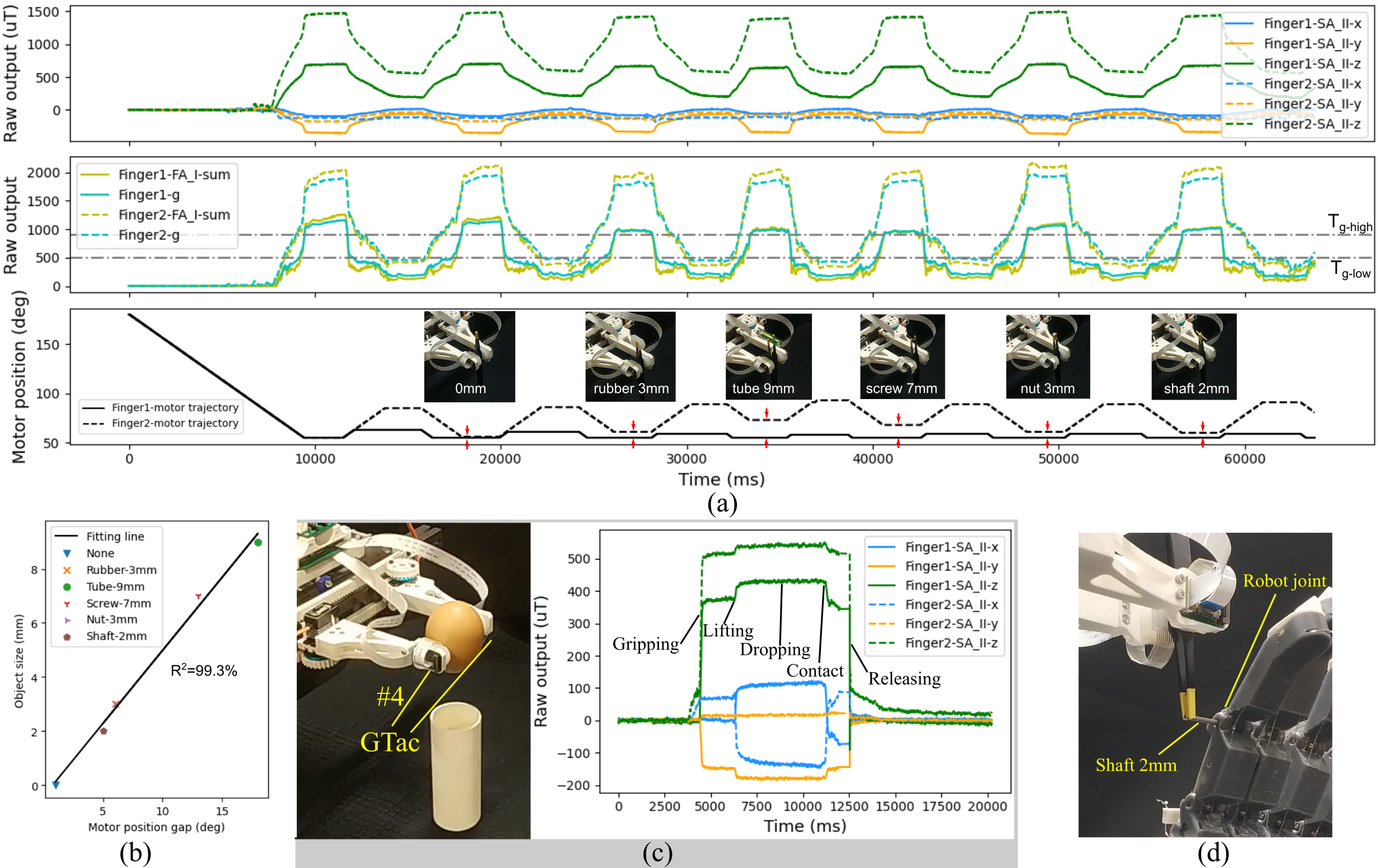}}
\caption{(a) Force-closed grasping for tool usage with the two-fingered gripper and the robotic arm. The raw force feedback from GTac on the gripper that was using tweezers for picking up small objects of various sizes. (b) The curve of object sizes versus motor position gap. (c) Left: the experiment photograph of the moment when the gripper grasped an egg in force-closed manner; Right: the raw feedback from SA-II layer of GTac on each fingertip where the key events were clearly distinguishable. (d) Experiment photographs of the process that the gripper was assembling the shaft into the finger joint of a robotic hand.}
\label{fig_gripper}
\end{figure*}

\subsection{Robotic integration}
\subsubsection{Delicate object grasping and tool usage using GTac enhanced gripper}
To demonstrate the force sensing capabilities of GTac in robotic tasks, experiments were conducted by a two-fingered gripper integrated with GTac (Fig. \ref{fig_gripper}(a)). Tweezers were used by the gripper for picking up small objects of various sizes in force-closed manner (Fig. \ref{fig_gripper}(a)). Since the contact force exerted on the objects by the tweezers is conducted by the gripper, the gap between two finger motor positions can be reflected as the fingertip displacement applied to squeeze the tweezers according to our force-closed controller. Therefore, the consistency and linearity between the motor gaps and the objects sizes suggest how controllable of the clipping force using tweezers which can be easily performed by humans. As our results shown (Fig. \ref{fig_gripper}(b)), the motors position gaps were linear to the various object sizes along a fitting line ($R^2=99.3\%$). Moreover, egg grasping and releasing experiments were performed to show the delicate object grasping and interactive force sensing capabilities of GTac-enhanced robots (Fig. \ref{fig_gripper}(c). According to the tactile feedback from the SA-II layer of GTac, the key contact event are clear distinguishable. Further, this tweezers usage capabilities were applied to assemble the shaft into the finger joint of a robotic hand so that the finger joint could rotate as normal (Fig. \ref{fig_gripper}(d)).

\section{Discussion}
Building upon previous researches on artificial tactile sensors and inspired by biological skin-like heterogeneity and integrability, we proposed a biomimetic tactile sensor, GTac, that can perceive naturally decoupled distributed normal and gross shear contact forces and provide the contact location and analogous joint-level torque estimation like the tactile sensing of human mechanoreceptors (FA-I and SA-II) can do. Our results showed that the human-like tactile sensing capabilities of GTac that mimicked human cutaneous structures enable robots to obtain heterogeneous force feedback and grasp fragile objects.

GTac can simultaneously provide normal and shear force sensing. Shear force sensing was achieved by measuring the change in the local magnetic flux density using a Hall sensor. Owing to the existence of the earth’s magnetic field and adjacent magnets, shear force sensing is disturbed when the orientation is altered, such as during finger flexion. According to our observations and experiments, the disturbance from the earth’s field can be reduced by 69.8\% through calibrating and estimating the current orientation using an IMU because the earth’s field is a constant field. In future work, it is important to shield GTac-enhanced robots from magnetic disturbances to achieve precise human-like tactile sensing capabilities during dynamic motion. More effective algorithms, such as ones that integrate visual feedback, could be developed to cancel such signal disturbances.

Regarding the skin-inspired design of GTac, the FA-I layer (piezoresistive sensors, extrinsic) and SA-II layer (Hall sensors, intrinsic) can be extended or scaled up separately for different purposes such as larger area sensing in the body of a humanoid \cite{mittendorferHumanoidMultimodalTactilesensing2011}, or prosthetics \cite{raspopovicSensoryFeedbackLimb2021a}. Tactile sensing characteristics, such as sensitivity, force range, and spatial resolution, can be customized by using various piezoresistive laminates (FA-I layer) or magnets of various strengths (SA-II layer) or silicon elastomer layers of various thicknesses and Young’s modulus. Therefore, GTac has the advantage of being able to customize both layers (Fig. \ref{fig_fabrication_custm}(b)) to obtain diversified tactile sensing characteristics for different functional regions that require them, such as hands and arms. GTac can also be feasibly used in soft robots by replacing the rigid PCB for acquiring tactile sensing signals with flexible printed circuits that are widely available. Therefore, the scalability and integrability of GTac can potentially render the human-like tactile sensing capabilities of GTac to various types of robots.

GTac is a hybrid solution of tactile sensor design that is uniquely suitable for integration into anthropomorphic robotic hands, which require dense normal and shear force sensing capabilities in a confined space, for acquiring human-like capabilities. Each finger of human hands consists of the distal, intermediate, and proximal phalanges and is connected to the palm in parallel. The tactile mechanoreceptors in human hands are primarily distributed at the finger and palm. Our ongoing work is to integrate GTac into the distal and proximal finger sections and the palm of an anthropomorphic robotic hand. The tactile sensors are connected serially in each finger (Fig. \ref{fig_fabrication_custm}(b)-(d)). The robotic hand would perceive 19 tactile sensing signals (16 from the FA-I layer and 3 from the SA-II layer) in each finger section and 285 tactile sensing signals (240 from the FA-I layer and 45 from the SA-II layer) in total.

\section{CONCLUSION}
\label{sec:conclusion}
In summary, we present a biomimetic tactile sensor that combines piezoresistive sensors and a Hall sensor to achieve simultaneous normal and shear force sensing. Due to its integrability, we integrated GTac into the fingertips of an robotic gripper and demonstrated that our sensor enabled the robotic gripper to acquire human skills, such as the grasping delicate objects based on its human-like sensing capabilities, illustrating its potential for various robotic applications, such as robotic grasping and manipulation, human-robot interaction, and the use of social robots.

\appendices
The supplementary video can be found in https://youtu.be/Pmd8PvLpeUA

\section*{Acknowledgment}
We thank H. Guo for assistance in execution of experiments, S. Bhattacharya, C.H. Low and D. Carmona for advising and assistance in mechatronics design and fabrication.

\printbibliography

\end{document}